\documentclass[runningheads]{llncs}
\usepackage[T1]{fontenc}
\usepackage{graphicx}
\usepackage{booktabs}
\usepackage[misc]{ifsym}

\usepackage{amsfonts}
\usepackage{amsmath}
\usepackage{subcaption}
\usepackage{multirow}
\usepackage{makecell}
\usepackage{pifont}
\usepackage{float}
\usepackage{enumitem}
\usepackage{microtype}
\begin{document}

\title{Collaborative Streaming Anomaly Detection with Interactive Explanations and Ensemble Consensus}

\titlerunning{Collaborative Streaming Anomaly Detection}

\author{Diogo Risca\inst{1} \and Afonso Lourenço\inst{1} \and Ricardo Martins\inst{2} \and Goreti Marreiros\inst{1}}

\authorrunning{Risca et al.}

\institute{GECAD, ISEP, Polytechnic of Porto, Portugal \email{\{difri, fonso, mgt\}@isep.ipp.pt} \and SISTRADE Software Consulting, Porto, Portugal \email{ricardo.martins@sistrade.com}}

\maketitle    

\begin{abstract}
We present a collaborative streaming anomaly detection system for high-speed data streams that explicitly integrates human analysts into the decision loop. The system combines heterogeneous detectors and aggregates their outputs through a normalization-based weighted consensus, complemented by artifact-aware rules to stabilize anomaly scoring under deployment. To improve interpretability, it derives surrogate models that approximate the ensemble consensus and expose human-readable sensor conditions associated with anomalous behavior. Analysts can actively intervene by reviewing anomaly episodes, adjusting consensus behavior, and refining surrogate rules used for anomaly prediction, producing a human-adjusted ensemble. We evaluate the approach on an industrial stream with 260\,000 events and 3 anomalous episodes, showing robust detection and actionable human-AI interaction.
\end{abstract}


\section{Introduction}

\begin{figure}[t]
    \centering
    \includegraphics[width=0.95\textwidth]{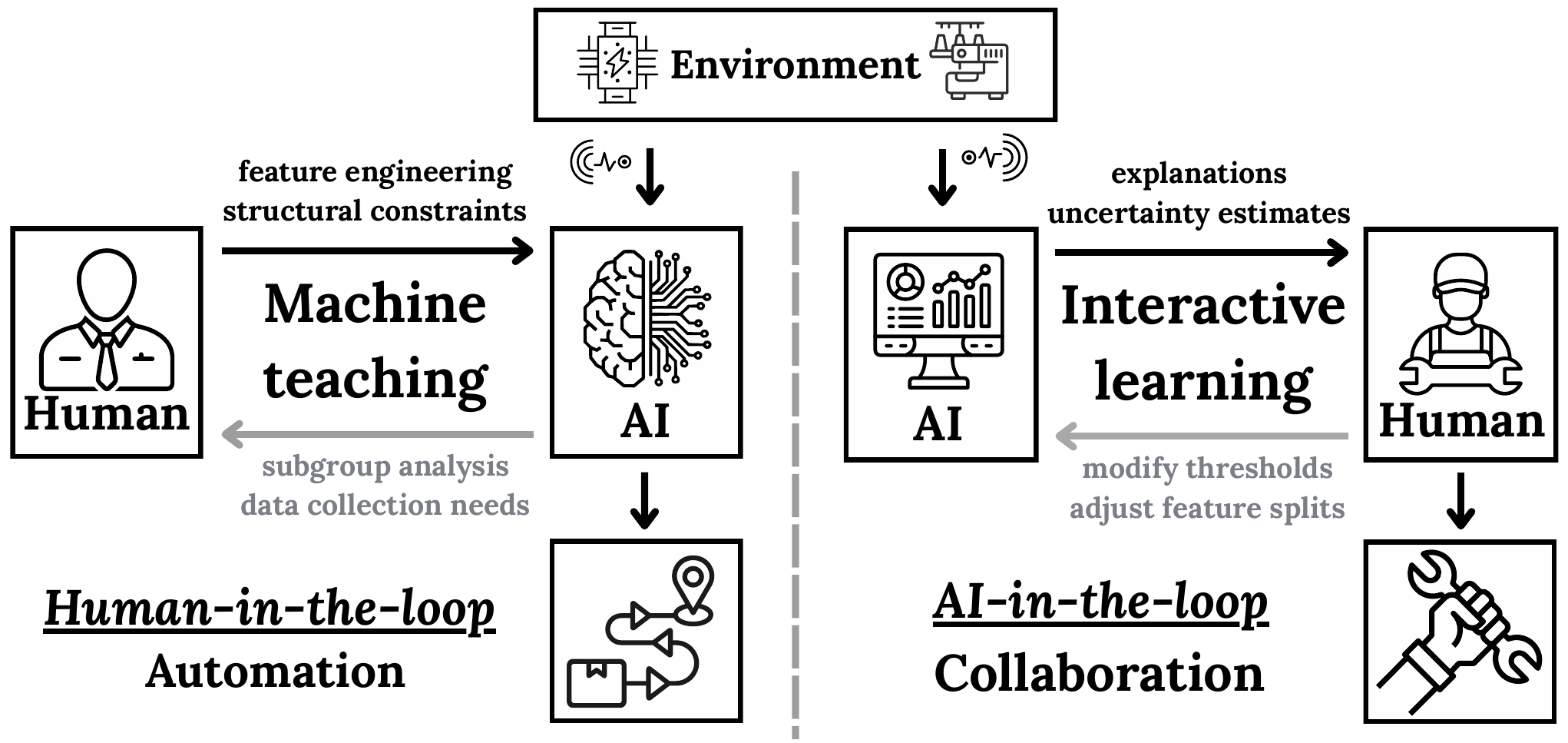}
    \caption{Human-AI interaction:  human-in-the-loop vs. AI-in-the-loop.}
    \label{fig:types}
\end{figure}

Real-time processing of high-speed data streams poses challenges due to unknown, evolving dynamics \cite{lourencco2025dfdt,lourencco2026context}. Streaming machine learning (SML) methods have been proposed, although most rely on fully supervised adaptation  \cite{neves2025online,lourencco2026ihomer+}. However, in real-world scenarios, it is unrealistic to provide ground-truth labels immediately upon prediction \cite{neves2026pitfalls,risca2025boosting,risca2025continual}. To address this, a wide variety of techniques have been proposed, with many mature open-source projects \cite{lourencco2025bridging,lourencco2025device}. However, a key question remains: what prevents their widespread adoption in real-world streaming systems? We argue that the challenge is building a technological stack that increases supervision in streams, ensuring that all methods operate effectively under deployment conditions. In this regard, increasing supervision naturally requires deliberate user effort, domain knowledge, and system-level support. In practice, formulating such design choices depends on whether an SML system is intended to operate in a fully automated or a collaborative setting \cite{natarajan2025human} (Fig. \ref{fig:types}).

When considering SML automation, the system shifts toward a machine teaching (MT) framework. In this human-in-the-loop regime, SML systems act autonomously on the environment while humans provide the necessary supervision. Information flows bidirectionally: humans provide models with labels, advice, constraints, or feature design, while models return diagnostics to humans that expose knowledge gaps, subgroup behavior, structural inconsistencies, and specific data requirements to better guide ongoing supervision \cite{ramos2020interactive,lourencco2026axle,crista2026streaming}. 

Conversely, SML collaboration establishes an interactive learning (IL) paradigm. In this AI-in-the-loop regime, SML systems support human decision-making rather than acting directly on the environment. Here, the information flow from models to humans manifests as explanations regarding decision rationale and explicit uncertainty estimates to signal what the model does not know. In response, humans refine the models by adjusting decision thresholds and feature representations based on these feedback signals \cite{sarailidis2023integrating,pauperio2025explainable}.

Grounded in this interactive learning paradigm, this work introduces a concrete collaborative SML system for high-speed industrial anomaly detection, integrating human analysts throughout the detection-to-action cycle via four tightly coupled modules: (i) a \emph{heterogeneous anomaly detection} module combining four streaming detectors with complementary structural assumptions \cite{manzoor2018xstream,sathe2016subspace,tan2011fast,guha2016robust}; (ii) a \emph{consensus ensembling} module that aggregates their scores through incremental normalization and truth discovery, with optional detector- and instance-level filtering \cite{yang2023foor,perini2023estimating,rayana2016less}; (iii) a \emph{rule-based surrogate modeling} module that explains the consensus in terms of human-readable sensor conditions, flags inconsistent score vectors for filtering, and monitors concept drift \cite{duarte2016adaptive}; and (iv) a \emph{human-AI interaction} module through which analysts filter spurious anomalies, adjust aggregation weights, select active detectors, and refine the rules used for anomaly prediction. The main contributions of this work are: (i) the formulation of a unified framework that operationalizes the collaborative interactive learning paradigm for streaming environments; (ii) the design of a multi-strategy consensus ensembling and rule-based surrogate modeling architecture that stabilizes and explains heterogeneous anomaly scores in real-time; and (iii) the practical validation through an industrial case study on a Jacquard loom.

\section{Case study}

\begin{table*}[t]
  \centering
  \caption{Description of the electrical features monitored on the Jacquard loom.}
  \label{tab:features}
  \begin{tabular}{m{2.3cm} @{\hspace{0.4cm}} m{5.7cm} @{\hspace{0.4cm}} m{3.5cm}}
    \toprule
    \textbf{Feature} & \textbf{Description} & \textbf{Fault Type}\\
    \midrule

    \makecell{Temperature} 
    & Indicates thermal stress 
    & Insulation degradation \\

    \makecell{NeutralCurrent} 
    & Indicates potential load imbalance 
    & Distribution faults \\

    \makecell{L1, L2, L3 \\ ActivePower} 
    & Measures average real power across phases to detect load distribution issues 
    & Distribution faults, phase imbalance \\

    \makecell{L1, L2, L3 \\ Current} 
    & Measures average current across phases to identify demand or fault signals 
    & Shorts, wiring faults, phase imbalance \\

    \makecell{L1, L2, L3 \\ ApparentPower} 
    & Summarizes total power across phases for overall system capacity insight 
    & System capacity issues \\

    \makecell{L1, L2, L3 \\ ReactivePower} 
    & Captures reactive power trends across phases for power factor analysis 
    & Capacitor/inductor faults \\

    \makecell{L1, L2, L3 \\ VoltageTHD} 
    & Assesses voltage harmonic distortion across phases for power quality 
    & Non-linear load degradation, harmonic issues \\

    \makecell{L1, L2, L3 \\ CurrentTHD} 
    & Assesses range of current harmonic distortion across phases for power quality 
    & Non-linear load degradation, harmonic issues \\

    \makecell{L1, L2, L3 \\ PhaseVoltage} 
    & Measures average voltage across phases to detect stability issues 
    & Voltage drops/spikes, system instability \\

    \makecell{L1, L2, L3 \\ PowerFactor} 
    & Summarizes power factor across phases for efficiency analysis 
    & Mechanical inefficiencies \\

    \bottomrule
  \end{tabular}
\end{table*}

This case study presents preliminary results from applying the proposed collaborative ensemble anomaly detection system to quality process control and predictive maintenance in a Jacquard loom, a specialized weaving machine that produces intricate patterns by individually controlling each warp thread, enabling applications such as automotive labels, logos, and decorative textiles. Characterizing performance degradation in this equipment requires considering its key electromechanical subsystems and their interdependencies. The primary 12\,kW motor drives weft insertion, warp thread actuation, and fabric winding, constituting the main source of mechanical motion and electrical load. The Jacquard head, mounted above the loom, actuates individual warp threads through solenoids or servomotors and is particularly susceptible to thermal stress, directly impacting pattern accuracy and machine availability. A secondary roller motor regulates fabric feed rate and weft density, introducing variable loads that propagate upstream and accelerate wear in coupled components.

To assess the health of the loom, the monitoring system extracts electrical indicators capturing both electrical and electromechanical degradation mechanisms, summarized with their targeted fault types in Table~\ref{tab:features}. The specific root causes of these anomalies are anonymized due to confidentiality constraints imposed by the industrial partner. The observation period comprises around 260\,000 monitored events, with three anomalous episodes.

\section{Methodology}

The proposed collaborative streaming framework is structured as a modular, four-stage pipeline that coordinates heterogeneous anomaly detection, consensus ensembling, rule-based surrogate modeling, and human-AI interaction.

\subsection{Anomaly detection}

Different anomaly detectors rely on distinct and often contrasting assumptions about the structure of abnormal events, including statistical, distance-based, density-based, and isolation-based principles. To capture complementary assumptions, we adopt four representative streaming anomaly detectors: \textit{xStream}, \textit{RSHash}, \textit{HST}, and \textit{RRCF}. \textit{xStream} \cite{manzoor2018xstream} estimates anomaly scores via multi-scale density approximation using sparse random projections and recursive half-space chains. By operating in projected spaces and aggregating across multiple resolutions, it reduces sensitivity to a single neighborhood scale and fixed subspaces. \textit{RSHash}~\cite{sathe2016subspace}, in contrast, avoids projections and assumes all original features contribute equally. It constructs ensembles of histograms over randomly selected feature subspaces, identifying anomalies as points falling into sparsely populated bins, offering a simpler density model than \textit{xStream}. \textit{HST}~\cite{tan2011fast} uses ensembles of random binary trees that rank anomalies by node mass rather than isolation depth. While computationally efficient, its uniform random splits yield coarser discrimination in high-dimensional spaces. \textit{RRCF}~\cite{guha2016robust} refines tree-based isolation by biasing splits toward high-variance features and defining anomalies by the structural disruption caused when inserting a point. Unlike \textit{HST}, it emphasizes localized deviations and supports anytime detection through incremental updates and forgetting.

\subsection{Consensus ensembling}

Anomaly detection is inherently unsupervised, which complicates reliability, threshold calibration, and interpretability. Practitioners often need to combine scores from heterogeneous detectors and guide decisions via aggregation, filtering, or feature selection. While evaluation metrics such as Mass-Volume, Excess-Mass, clustering indices, or classifier separability can assess detector quality, applying them to high-dimensional data requires preprocessing and ensembling strategies similar to detection itself, essentially recreating the original reliability challenge. To address this, we frame anomaly detection ensembling as a two-stage process: normalization and truth discovery. Central to this framework is a reference score matrix $S \in \mathbb{R}^{n \times m}$ of anomaly scores, where $n$ is the number of instances and $m$ is the number of detectors. Each stage of the ensembling pipeline operates on this matrix, yielding 12 possible strategies:
\[
\underbrace{
    \underbrace{1}_{\text{Z-Score}}
}_{\text{Normalization}}
\times 
\underbrace{
    \underbrace{2}_{\text{Yes / No}}
}_{\text{Instance filtering}}
\times
\underbrace{
    \underbrace{3}_{\text{Full / DivE / ULARA}}
}_{\text{Truth Discovery}}
\times 
\underbrace{
    \underbrace{2}_{\text{Yes / No}}
}_{\text{Detector filtering}}
= 12
\]

\begin{figure}[h]
    \centering
    \includegraphics[width=\textwidth]{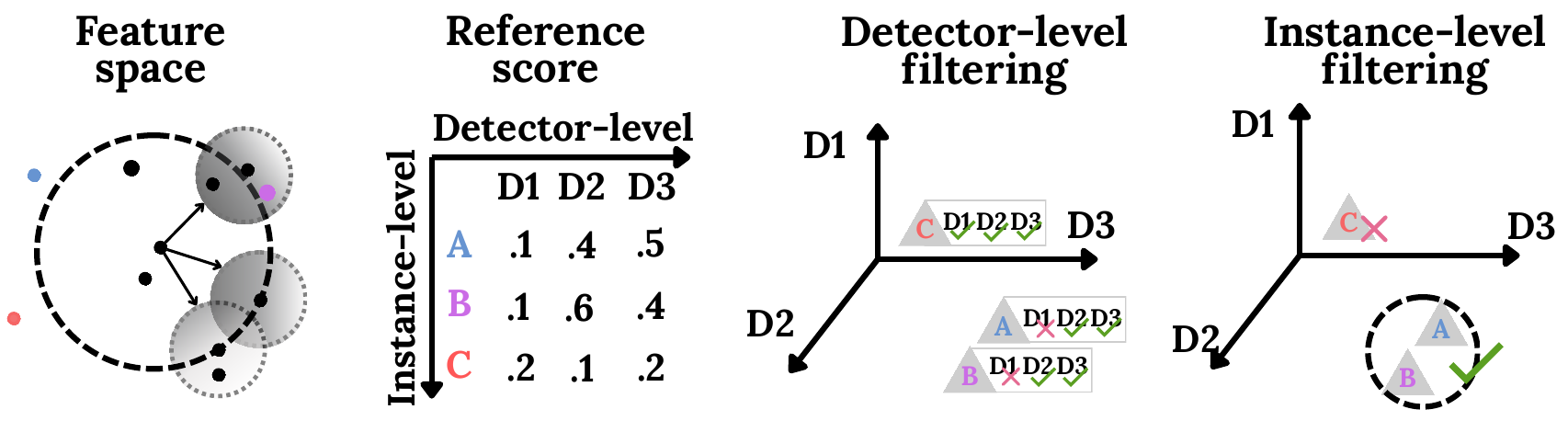}
    \caption{Instance- and detector-level filtering}
    \label{fig:foor}
\end{figure}

\textit{Normalization} converts detector scores in $S$ into comparable probabilities \cite{kriegel2011interpreting}. Each column $S'_{:,j}$ is 0--1 normalized using an incremental Z-Score calibrator. Normalization can operate globally over all historical scores or within a rolling window. Optionally, we add instance-level filtering (Figure \ref{fig:foor}). Following methods like FOOR \cite{yang2023foor} and Perini \cite{perini2023estimating}, we analyze each row $S_{i,:}$ to identify inconsistent score vectors as those not covered by any AMRule \cite{duarte2016adaptive} (details in Section~\ref{sec:rbsm}). $S'$ denotes the filtered reference matrix.

\textit{Truth Discovery} combines columns of $S'$ to obtain a single score $y$ per instance, with three optional strategies: Full: Compute the row-wise average of $S'$, i.e., $y_i = \frac{1}{m} \sum_{j=1}^m S'_{i,j}$; DivE: Select a subset of columns $S'_{:,J} \subseteq S'$ that are maximally diverse, then average or weight them; and ULARA: Assign weights $w_j$ to each column based on agreement with other columns on top-$k$ anomalies, producing $y_i = \sum_{j=1}^m w_j S'_{i,j}$ \cite{rayana2016less}. Optionally, we add detector-level filtering. Following methods like LSCP \cite{zhao2019lscp}, and AutoOD \cite{cao2023autood}, columns $S'_{:,j}$ that poorly align with a pseudo-ground truth of $y$ are incrementally removed.

\subsection{Rule-based surrogate modeling}
\label{sec:rbsm}

With an ensemble of heterogeneous anomaly detectors, a key challenge is to explain why certain instances are flagged as anomalous. To address this, one can both construct a \emph{raw-data surrogate} that predicts the consensus anomaly probability directly from the original features, providing explanations as attribute-specific intervals for a small subset of the feature space. Second, as a \emph{consensus-score surrogate}, it predicts the consensus anomaly from the outputs of multiple models, offering interpretable insights into how individual anomaly scores combine. Following methods such as STAIR \cite{deng2024outlier}, and PARs \cite{feng2024pars}, we adopt rule-based surrogates. In particular, we adopt AMRules \cite{duarte2016adaptive} to incrementally build a set of rules. Each rule $r \in RS$ maintains sufficient statistics of the examples it covers, and a default rule handles uncovered instances. For each example $(\mathbf{x},y)$, rules covering it update their statistics. The \textit{Page-Hinkley (PH) test} monitors the prediction error; if a significant change is detected, the rule is removed. Rules expand after $N_\text{min}$ examples using the attribute with the highest \textit{variance ratio}:
\begin{equation}
\text{VR}(h_A) = 1 - \frac{|E_L|}{|E|} \frac{\mathrm{var}(E_L)}{\mathrm{var}(E)} - \frac{|E_R|}{|E|} \frac{\mathrm{var}(E_R)}{\mathrm{var}(E)},
\end{equation}
where $E$ is the set of examples seen by the rule, and $E_L, E_R$ are partitions by a potential split. Hoeffding bounds and a threshold $\tau$ ensure statistically reliable splits. Numeric attributes use a limited \textit{Extended Binary Search Tree} for efficient split management. Continuous attributes use Gaussian or Cantelli-based probability estimates. Each rule predicts using an adaptive choice between the mean of covered targets or an incremental linear model based on lowest MAE with fading factor $\alpha$. Rules can be \textit{ordered} (first covering rule predicts) or \textit{unordered} (all rules contribute, averaged).

\begin{figure}[t]
    \centering
    \begin{minipage}[t]{0.56\textwidth}
        \centering
        \includegraphics[height=5.2cm, keepaspectratio]{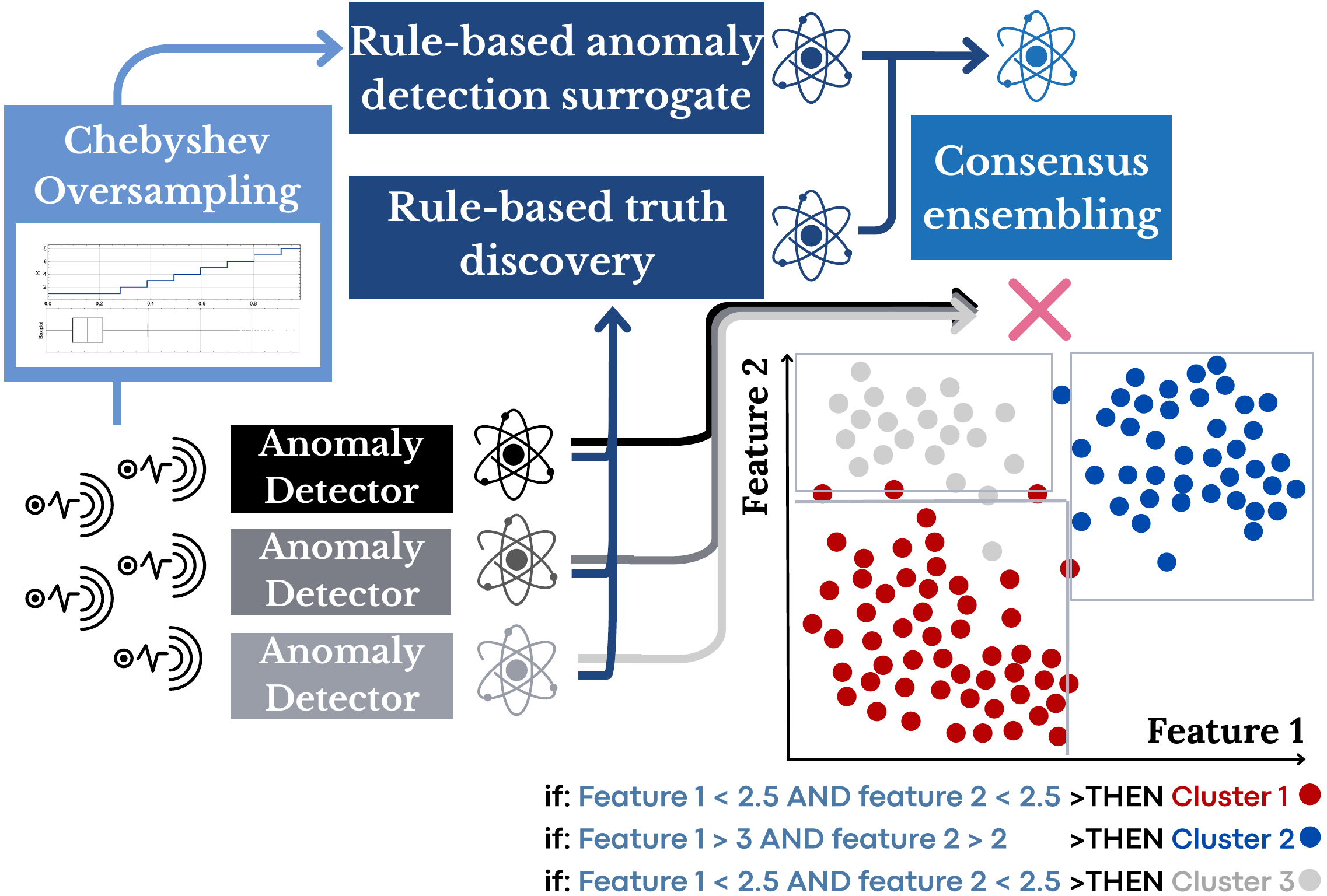}
        \caption{Rule-based surrogate.}
        \label{fig:surrogate}
    \end{minipage}%
    \hfill
    \begin{minipage}[t]{0.40\textwidth}
        \centering
        \includegraphics[height=5.2cm, keepaspectratio]{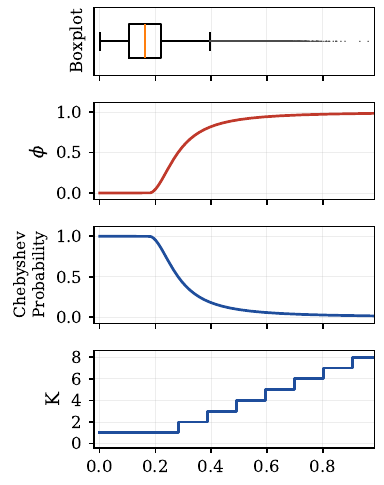}
        \caption{Chebyshev.}
        \label{fig:cheby}
    \end{minipage}
\end{figure}

To improve rule-based anomaly detection surrogate learning under highly imbalanced reconstruction error distributions, we adopt the ChebyOS oversampling strategy \cite{aminian2021chebyshev}. The approach exploits Chebyshev's inequality to estimate the rarity of an observation according to its distance from the mean. Given the running mean $\bar{y}$ and standard deviation $\sigma$ of the reconstruction error, the normalized distance of an incoming example is computed as $t=\frac{|y-\bar{y}|}{\sigma}.$ Chebyshev's inequality states that:
\begin{equation}
P(|Y-\bar{y}| \geq t\sigma)\leq\frac{1}{t^2},
\end{equation}
indicating that observations farther from the mean become increasingly unlikely. Since high scores are more likely to correspond to anomalous operating conditions, these examples are emphasized during learning by presenting each incoming observation to the learner $K=\left\lceil\frac{|y-\bar{y}|}{\sigma}\right\rceil$ times.

In the consensus-score surrogate of black-box anomaly scores, rule statistics also allow to detect outlier-score outliers:
\begin{equation}
\text{Ascore} = \frac{1}{d} \sum_{j=1}^{d} \log \frac{P(X_j=v|L_r)}{1-P(X_j=v|L_r)},
\end{equation}
where $L_r$ stores sufficient statistics for rule $r$. Examples with \textit{Ascore} below threshold $t$ are flagged as inconsistent score vectors for the instance-level filtering (described in previous section).

\subsection{Human-AI interaction}

\begin{figure}[h]
    \centering
    \includegraphics[width=\textwidth]{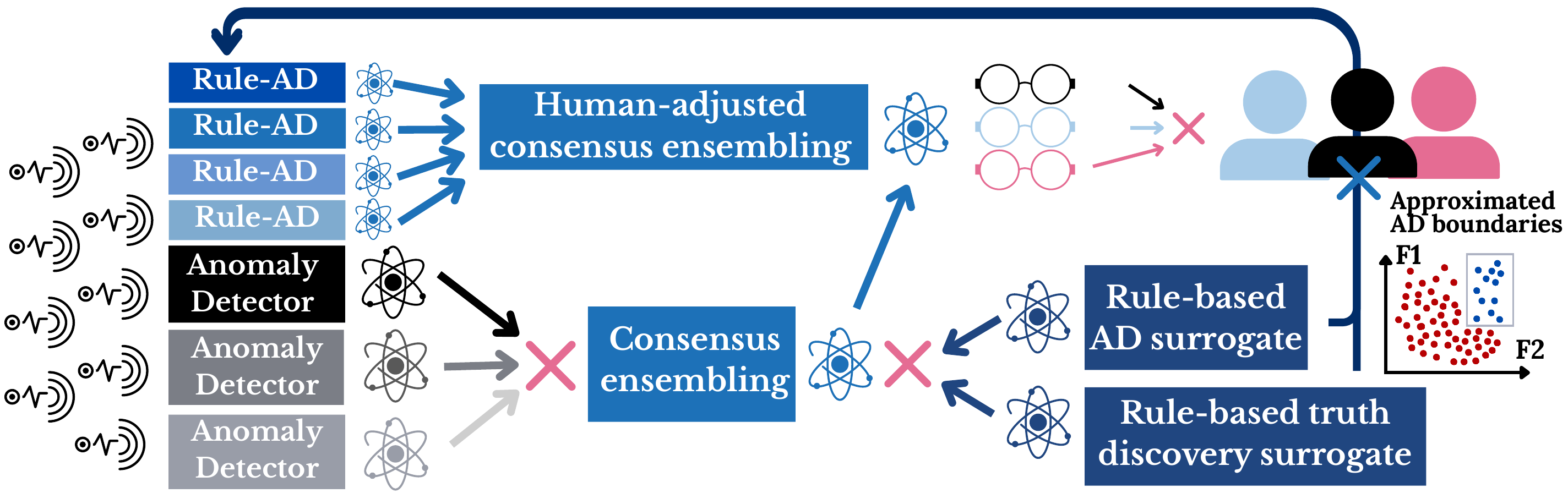}
    \caption{Human-adjusted consensus ensembling}
    \label{fig:adjustement}
\end{figure}

In this AI-in-the-loop regime, all stages act as information flows from the model to humans through explanations (“I know why I decided this”) and uncertainty estimates (“I know what I don’t know”). These signals not only support decision-making, but also enable humans to actively intervene by adjusting decision thresholds and feature representations. Concretely, the framework allows domain experts to inject knowledge by filtering spurious anomalies (e.g., known production regime changes), adjusting aggregation weights, selecting subsets of detectors, and choosing the most consistent and interpretable surrogate rules for inference (Figure~\ref{fig:adjustement}). This results in a human-adjusted consensus ensemble, where outputs are refined rather than merely consumed.

\section{Results}

Before evaluating the proposed ensemble framework, we first analyze the behavior of the individual anomaly detectors throughout the operational cycle of the Jacquard loom. During normal operation, the machine alternates between active weaving periods and extended idle states, during which electrical measurements converge toward near-baseline values. These operating mode transitions shift the data distribution without any relation to machine faults, so unsupervised detectors may flag them as anomalous, generating many false positives. To mitigate this, a deterministic artifact-filtering rule was introduced to model the sensor signatures of idle operation, identifying zero-load conditions and suppressing anomaly alerts during these periods.

\begin{figure}[h]
\centering
\includegraphics[width=\textwidth]{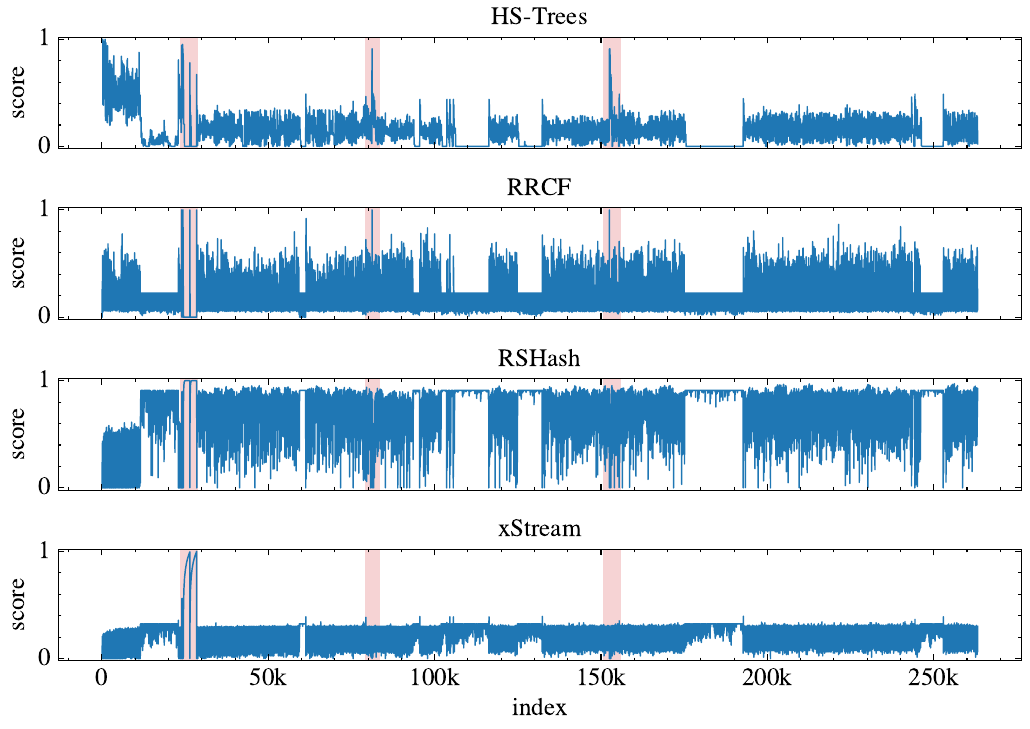}
\caption{Normalized anomaly scores generated by HS-Trees, RRCF, RSHash, and xStream across the stream. Red shaded regions indicate anomaly intervals.}
\label{fig:detector_scores}
\end{figure}

After accounting for idle-state artifacts, the four streaming anomaly detectors exhibit markedly different performance when evaluated independently against the labeled anomaly events, with ROC-AUC scores ranging from 97.71 for HS-Trees to 45.77, 11.64, and 7.64 for RRCF, RSHash, and xStream, respectively. Figure~\ref{fig:detector_scores} presents the normalized anomaly scores produced by each detector alongside the ground-truth anomaly intervals. While some methods respond strongly to normal operational fluctuations, others fail to capture the degradation patterns of confirmed anomalies, showing that no single detector is consistently reliable across all operating conditions and motivating a consensus-based ensemble approach.

\begin{figure}[h]
\centering
\includegraphics[width=\textwidth]{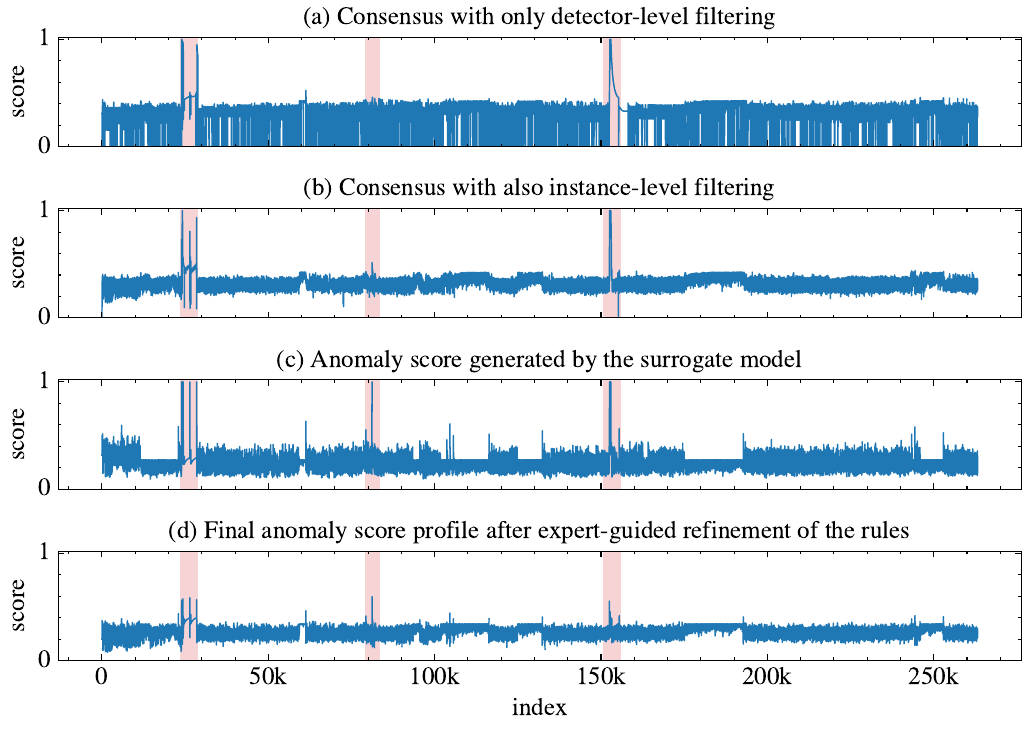}
\caption{Evolution of the anomaly detection pipeline.}
\label{fig:anomaly_pipeline}
\end{figure}

\begin{figure}[h]
\centering
\includegraphics[width=\textwidth]{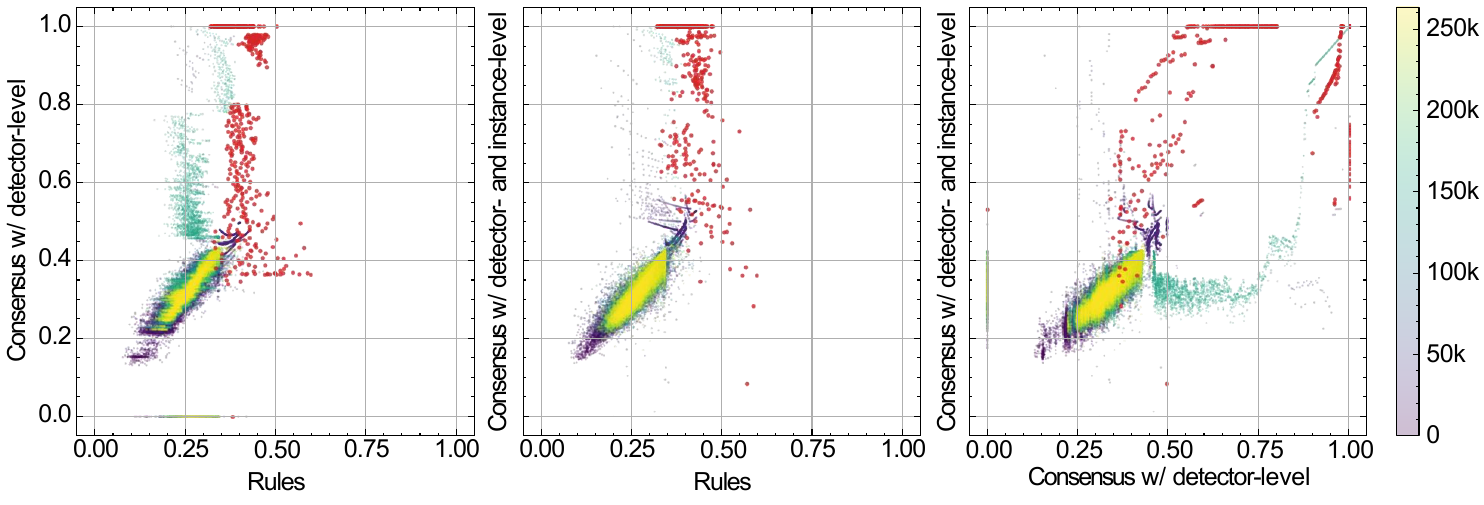}
\caption{Interaction across detector-, instance-, and expert-level components.}
\label{fig:ad}
\end{figure}

To integrate the complementary information provided by the individual detectors, the anomaly score matrix $S \in \mathbb{R}^{n \times m}$ is processed through the proposed consensus ensembling framework. Figure~\ref{fig:anomaly_pipeline}(a) shows the profile produced with incremental Z-score normalization and detector-level filtering, achieving a ROC-AUC of 99.08. Figure~\ref{fig:anomaly_pipeline}(b) presents the complete framework, which adds instance-level filtering with rule-coverage validation to remove inconsistent detector outputs and transient statistical artifacts prior to truth discovery, reaching a ROC-AUC of 99.46. As shown in Figure~\ref{fig:ad}, this filtering reduces background score variability and eliminates many ambiguous mid-range scores, improving the separation between normal operation and confirmed anomalies. Following consensus generation, a surrogate model was trained directly on the raw sensor features and, as shown in Figure~\ref{fig:anomaly_pipeline}(c), reproduces the principal anomaly patterns identified by the ensemble while operating exclusively in the original feature space. Finally, the extracted surrogate rules were reviewed by domain experts: rules associated with known operational conditions or production adjustments were removed, while intervals consistent with engineering knowledge were retained and refined. The resulting human-adjusted profile (Figure~\ref{fig:anomaly_pipeline}(d)) reduces residual background fluctuations, improves interpretability, and recovers anomalies that were previously masked.

\begin{figure}[h]
    \centering
    \includegraphics[width=\textwidth]{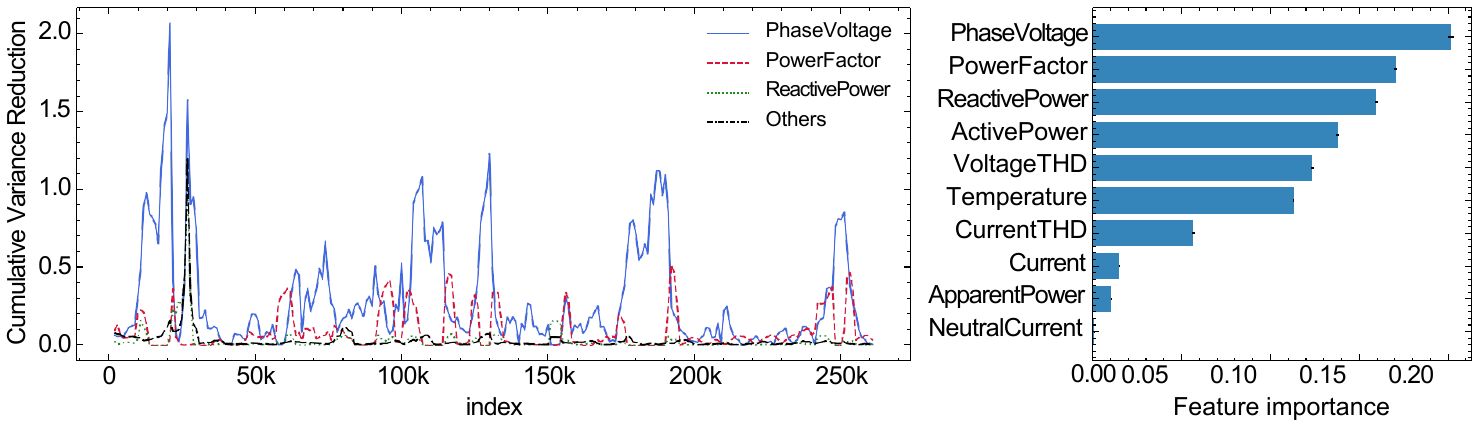}
    \caption{Feature importance from rule split quality via variance reduction.}
    \label{fig:pfi_group_comparison}
\end{figure}

Figure~\ref{fig:pfi_group_comparison} shows that PhaseVoltage, PowerFactor, and ReactivePower consistently emerge as the most influential variables. These features are directly related to electrical efficiency, load balance, and power quality, suggesting that the detected anomalies are primarily associated with changes in the electromechanical operating conditions of the loom rather than isolated sensor fluctuations.

The raw-data surrogate translates the ensemble decisions into two interpretable rule branches that were refined incrementally through human-AI interaction. After Episode~1, the domain expert reviewed the surrogate output and retained/refined $R_1$ to represent voltage collapse, where over-current coincides with a phase-voltage sag ($<0.85\times$ median), consistent with severe electrical loading or short-circuit conditions. After Episode~2, the newly observed waveform-distortion pattern led to the introduction of $R_2$, where over-current occurs while voltage remains above the collapse threshold but current harmonic distortion increases ($\text{maxCurrentTHD}>0$). For Episode~3, the existing rule set ${R_1,R_2}$ was sufficient to characterize the anomaly:
\begin{align*}
  R_1 &\equiv \text{maxCurrent} > \tau_I \land \text{minPhaseVoltage} < \tau_V \\
  R_2 &\equiv \text{maxCurrent} > \tau_I \land \text{minPhaseVoltage} \ge \tau_V \land \text{maxCurrentTHD} > 0 \\
  \text{Anomaly} &\equiv R_1 \lor R_2
\end{align*}
where $\tau_I=Q_{0.995}(\text{maxCurrent})$ and $\tau_V=0.85\cdot\text{median}(\text{minPhaseVoltage})$. Both rules share an over-current premise but diverge according to the observed electrical manifestation. Retrospectively, the resulting rule set characterized the observed anomaly patterns across all three episodes (Figure~\ref{fig:detector_scores}): Episode~1 is dominated by voltage collapse ($\sim89\%\ R_1$), whereas Episodes~2 ($\sim60\%\ R_1/\sim40\%\ R_2$) and 3 ($\sim54\%\ R_1/\sim46\%\ R_2$) combine both patterns. Since each intervention occurred only after the corresponding episode had been processed, subsequent rule refinements did not use information from future episodes.

Computational evaluation across the 260\,000 events stream reveals clear throughput trade-offs. Individual detector runtimes scaled as RRCF ($\sim 1080\,\text{s}$) $>$ HST ($\sim 480\,\text{s}$) $>$ xStream ($\sim 150\,\text{s}$) $>$ RSHash ($\sim 95\,\text{s}$). Regarding framework add-ons, the rule-based surrogate introduced the main overhead ($\sim 1600\,\text{s}$), whereas detector-level consensus ($\sim 45\,\text{s}$), instance-level filtering ($\sim 420\,\text{s}$), and expert rule application ($\sim 12\,\text{s}$) added minimal latency, preserving overall efficiency.

\section{Conclusions}

We presented a collaborative streaming anomaly detection framework that integrates heterogeneous detectors, consensus ensembling, rule-based surrogates, and human-in-the-loop refinement. Applied to a 260\,000 event industrial stream from a Jacquard loom, the framework improved detection performance from 97.71 ROC-AUC (best individual detector) to 99.46 ROC-AUC. Crucially, the raw-data surrogate distilled the ensemble decision boundaries into two physically interpretable rules that separate voltage-collapse from waveform-distortion faults across all three anomaly episodes. This provides domain engineers with transparent, domain-grounded thresholds that can be audited and recalibrated under new operating conditions. While validated on a single industrial setup, the architecture establishes a generalizable pipeline for actionable human-AI interaction in streaming environments. Future work will focus on integrating automated preference learning from analyst feedback.

\begin{credits}
\subsubsection{\ackname} Funded by the Portuguese Republic’s Recovery and Resilience Plan, under PRODUTECH R3, and FCT, under project 10.54499/UID/00760/2025.

\subsubsection{\discintname}
The authors have no competing interests to declare that are relevant to the content of this article.
\end{credits}

%
%
%
\bibliographystyle{splncs04}
\bibliography{references}
%

\end{document}